\documentclass[a4paper,UKenglish,cleveref, autoref, thm-restate]{lipics-v2021}

\usepackage{booktabs}
\usepackage{xcolor}
\usepackage[most]{tcolorbox}
\newtcolorbox{custom_box}{
    enhanced,
    colback=gray!8,
    colframe=gray,
    boxrule=0.8pt,
    arc=2mm,
    left=2mm,
    right=2mm,
    top=1mm,
    bottom=1mm,
    fonttitle=\bfseries,
    title=Key Considerations,
    coltitle=white
}
\usepackage{tikz}
\newcommand{\circled}[1]{%
\tikz[baseline=(char.base)]{
\node[shape=circle,draw,inner sep=1pt] (char) {#1};}}

\title{A Framework for the Systematic Review of ML Assets in AI Registries}

\author{Alexandra González}{Universitat Politècnica de Catalunya - BarcelonaTech (UPC), Barcelona, Spain}{alexandra.gonzalez.alvarez@upc.edu}{0009-0003-7634-0343}{}
\author{Quim Motger}{Universitat Politècnica de Catalunya - BarcelonaTech (UPC), Barcelona, Spain}{joaquim.motger@upc.edu@upc.edu}{0000-0002-4896-7515}{}
\author{Xavier Franch}{Universitat Politècnica de Catalunya - BarcelonaTech (UPC), Barcelona, Spain}{xavier.franch@upc.edu}{0000-0001-9733-8830}{}
\author{Silverio Martínez-Fernández}{Universitat Politècnica de Catalunya - BarcelonaTech (UPC), Barcelona, Spain}{silverio.martinez@upc.edu}{0000-0001-9928-133X}{}

\authorrunning{Anonymous Author(s)} 

\Copyright{Anonymous Author(s)} 

\ccsdesc[500]{General and reference~Surveys and overviews}
\ccsdesc[500]{Software and its engineering~Software creation and management}

\keywords{Models, Datasets, AI registries, Systematic Reviews}

\category{} 

\relatedversion{} 

\nolinenumbers 

\EventEditors{John Q. Open and Joan R. Access}
\EventNoEds{2}
\EventLongTitle{42nd Conference on Very Important Topics (CVIT 2016)}
\EventShortTitle{CVIT 2016}
\EventAcronym{CVIT}
\EventYear{2016}
\EventDate{December 24--27, 2016}
\EventLocation{Little Whinging, United Kingdom}
\EventLogo{}
\SeriesVolume{42}
\ArticleNo{23}

\begin{document}

\maketitle

\begin{abstract}
\textbf{Background:} Modern software systems increasingly rely on Machine Learning (ML) assets (i.e., pre-trained models, datasets, benchmarks) for building, evaluating, and integrating ML-based systems. However, current exploration, selection and reuse practices of ML assets are not supported by systematic retrieval methodologies comparable to those used in traditional evidence synthesis. Consequently, in practice, ML asset selection is often presented as a settled design decision, supported by informal justification rather than a traceable, evidence-based, and updatable selection process.
\textbf{Aims:} This paper explores how systematic review methods can support ML asset retrieval. In doing so, we aim to make their selection transparent and reproducible, grounded in explicit evidence, and ultimately better suited to its intended use.
\textbf{Method:} We analyze established systematic review practices from scientific literature and adapt their phases (i.e., planning, conducting, and documenting) to Artificial Intelligence (AI) registries, treating ML assets as first-class units of analysis. The resulting framework integrates registry-aware search strategies, cross-registry schema alignment, and dependency-driven ML asset exploration.
\textbf{Results:} We conceptualize ML asset retrieval as a systematic and reproducible process rather than an ad hoc activity, and propose a framework for structured ML asset discovery. 
\textbf{Conclusions:} This work illustrates how systematic review principles can be extended beyond scientific literature to support evidence synthesis over evolving AI registries.
\end{abstract}

\section{Introduction} 
Modern software development workflows increasingly rely on Machine Learning (ML) assets, including pre-trained models, datasets, and benchmarks \cite{zhao2024empirical, stalnaker2025empirical, ait2023hfcommunity}. In this context, researchers and practitioners use open, large-scale registries such as Hugging Face and Kaggle, often referred to as Artificial Intelligence (AI) ecosystems \cite{yang2025ecosystem}, to select models for system integration, reuse datasets in experiments, or identify benchmarks for evaluation. 

Despite this growing reliance on ML assets, current practices for selecting them remain ad hoc and are rarely documented in a systematic way. This lack of standardization reflects a fundamental mismatch between ML assets and traditional scholarly publications. Unlike research papers, ML assets are hosted in rapidly evolving registries with sparse documentation and volatile (if any) quality signals \cite{castano2024analyzing, oreamuno2024state, stalnaker2025empirical}. Although initiatives such as model cards \cite{mitchell2019model}, datasheets for datasets \cite{gebru2021datasheets}, and more recent efforts such as AI Bill of Materials \cite{10.1145/3786773} aim to improve transparency and standardize artifact documentation, they do not provide methodological guidance for systematically retrieving ML assets in AI registries. 
Furthermore, different registries expose heterogeneous schemes and traceability signals. For instance, registries such as Hugging Face provide popularity indicators (e.g., likes) and explicit traceability links to downstream models trained or fine-tuned on a given dataset. In contrast, others, such as Kaggle, rely on alternative metrics (e.g., upvotes) while offering limited support for tracing reuse. This leads to uneven levels of traceability across registries, where relationships between ML assets may be explicitly documented or only implicitly inferred. When considered jointly, these heterogeneous signals form a complex network, spanning links between datasets and models, model-to-model adaptations (e.g, quantization, fine-tuning) \cite{zhou2025unifying}, and dataset-to-dataset lineages (e.g., subsets, augmentations) \cite{koch2021reduced}. This network is critical, as a model is unlikely to perform reliably in production if its deployment context diverges from its training data, or if these datasets reflect unwanted biases \cite{gebru2021datasheets, pepe2024hugging}. 

These challenges reveal the need for more rigorous and reproducible methodologies for ML asset discovery. To address this need, systematic reviews and related evidence synthesis methodologies offer a promising foundation. 
They provide rigorous procedures to guide the identification, selection, and analysis of relevant studies, consolidating the state of the art within a research field. 
In Software Engineering (SE), secondary studies methodologies such as Systematic Literature Reviews (SLRs) \cite{kitchenham2007guidelines, brereton2007lessons}, Systematic Mapping Studies (SMSs) \cite{petersen2008systematic}, and Rapid Reviews \cite{cartaxo2020rapid} support different evidence synthesis objectives, ranging from in-depth analysis to broad topic exploration and accelerated review processes. 
These methodologies define structured processes for planning, conducting, and reporting reviews, while emphasizing transparency and reproducibility. They rely on explicit metadata (e.g., titles, abstracts, keywords) and stable indexing provided by traditional academic databases (e.g., Scopus, IEEE Xplore, ACM Digital Library), which together enable controlled and repeatable study retrieval. However, traditional evidence synthesis methods are not directly applicable to AI registries. While they assume structured and standardized corpora, AI ecosystems are dynamic, heterogeneous, and governed by evolving and partially implicit dependency relationships. These differences are summarized in Table~\ref{tab:comparison}. 

\begin{table}[!h]
    \centering
    \caption{Key differences between systematic reviews of scientific papers and ML assets.}
    \label{tab:comparison}
    \small
    \begin{tabular}{lll}
        \toprule
        & \multicolumn{2}{c}{\textbf{Systematic Reviews of...}} \\
        \cmidrule(lr){2-3}
        \textbf{Aspect} & \textbf{Scientific Papers} & \textbf{ML Assets} \\
         \midrule
         Primary unit & Academic publications & Models, datasets, benchmarks \\
         Data sources & Academic databases & AI registries \\
         Metadata quality & Structured and standardized & Sparse, inconsistent, heterogeneous \\
         Relationships & Citation networks & Model-dataset dependencies \\
         Reproducibility & Stable indexed records & Rapidly evolving registries and metadata \\
    \bottomrule
    \end{tabular}
\end{table}

This paper addresses this mismatch by proposing a framework for systematically retrieving ML assets from AI registries, inspired by systematic reviews of scientific literature and adapted to the characteristics of open, heterogeneous, and continuously evolving data sources.

\section{Related Work}
Empirical SE studies that rely on ML assets typically treat their retrieval as an implementation detail rather than a formalized research step. In practice, datasets, pre-trained models, and benchmarks are selected through ad hoc filtering strategies, manually curated subsets, or registry-specific heuristics that are rarely documented in a systematic form. For instance, Prenner et al. \cite{prenner2021making} evaluated pre-trained transformer models on 13 small SE datasets selected from the literature guided by task diversity, dataset size, and availability of baselines. Similarly, other studies adopt pragmatic selection criteria when constructing experimental corpora, although these vary across objectives. Durán et al. \cite{duran2025insights} selected 12 code generation models from Hugging Face by constraining model size, focusing on base models, and ensuring representativeness across sizes, release dates, and affiliations. González et al. \cite{gonzalez2025impact} further filtered models based on the availability of training dataset information, popularity signals, and size constraints. Across these works, ML asset inclusion is driven by experimental requirements rather than reproducible retrieval strategies grounded in registry structure.

A complementary line of work investigates AI ecosystems as interconnected networks of models, datasets, and related artifacts. These studies shift the focus from individual ML assets to their relationships, highlighting that ML artifacts form complex dependency structures. Dasoulas et al. \cite{dasoulas2024mlsea} proposed a knowledge graph integrating datasets, pipelines, implementations and scientific works from sources such as OpenML, Kaggle, and Papers with Code. Similarly, Yang et al. \cite{yang2025ecosystem} analyzed the ecosystem of Large Language Models (LLMs) for code in Hugging Face, combining manual and LLM-assisted analysis, and incorporating relationships between models and their training datasets through snowballing-based exploration. Jiang et al. \cite{10.1145/3643991.3644907} explored cross-repository organization of AI artifacts across Hugging Face and GitHub. From an SE perspective, cataloguing efforts have examined registries such as Hugging Face and highlighted inconsistencies in metadata quality \cite{gonzalez2025cataloguing}. While these works provide important insights into the structure of AI ecosystems, they primarily focus on analyzing them rather than using their structure to support systematic retrieval.

Taken together, these streams of work point to a dual reality: ML assets are central to empirical SE and its intersection with AI, and are embedded in complex ecosystems with rich dependency structures. However, evidence synthesis practices remain grounded in document-centric corpora and have not evolved accordingly. From the perspective of prior research on ML asset usage and AI ecosystem analysis, there is still no established approach. 
The following points summarize the most relevant implications of this lack of formalization:

\begin{itemize}
\item \textbf{Obscure selection rationale.} ML asset choices are rarely accompanied by explicit criteria, leaving the reasoning behind a given selection opaque and difficult to assess.
\item \textbf{Limited reproducibility.} Without documented selection criteria, verifying if an ML asset suits a specific context is difficult, hindering asset reuse and cross-study comparison.
\item \textbf{Suboptimal selection outcomes.} Without the guidance on where, what, and how to search, ML asset discovery becomes ad hoc, rarely yielding the best results for the task.
\item \textbf{Costly and unreliable re-execution.} Without documented retrieval processes, fast-evolving ML catalogs make repeating a data selection effortful, error-prone, and unreliable.
\end{itemize}

This gap exposes the need for methodological support that enables structured, transparent, and reproducible ML asset discovery in evolving AI ecosystems.

\section{The Review Process}
To formalize the derivation of the proposed framework, we adopted a structured design process rooted in established systematic practices and state-of-the-art literature. First, we analyzed protocols for systematic reviews, focusing on SLRs and their operational phases of planning, conducting, and documenting reviews \cite{kitchenham2007guidelines, brereton2007lessons}. Second, we conducted a thematic analysis of primary studies addressing ML asset selection, encompassing recent model-centric \cite{di2024automated, gonzalez2025cataloguing, yang2025ecosystem} and data-centric \cite{motger2025share} perspectives. This allowed us to isolate both their operational processes and recurring methodological challenges, such as metadata incompleteness, dependency-aware exploration, and reproducibility threats. Third, we explicitly aligned the processes and challenges extracted from these studies and the core SLR phases and steps. This synthesis directly shaped the dimensions of the resulting methodology, ensuring that these emerging results remain grounded while addressing the concrete, open challenges of evolving AI registries.

\begin{figure}[!htb]
    \centering
    \includegraphics[width=1\linewidth]{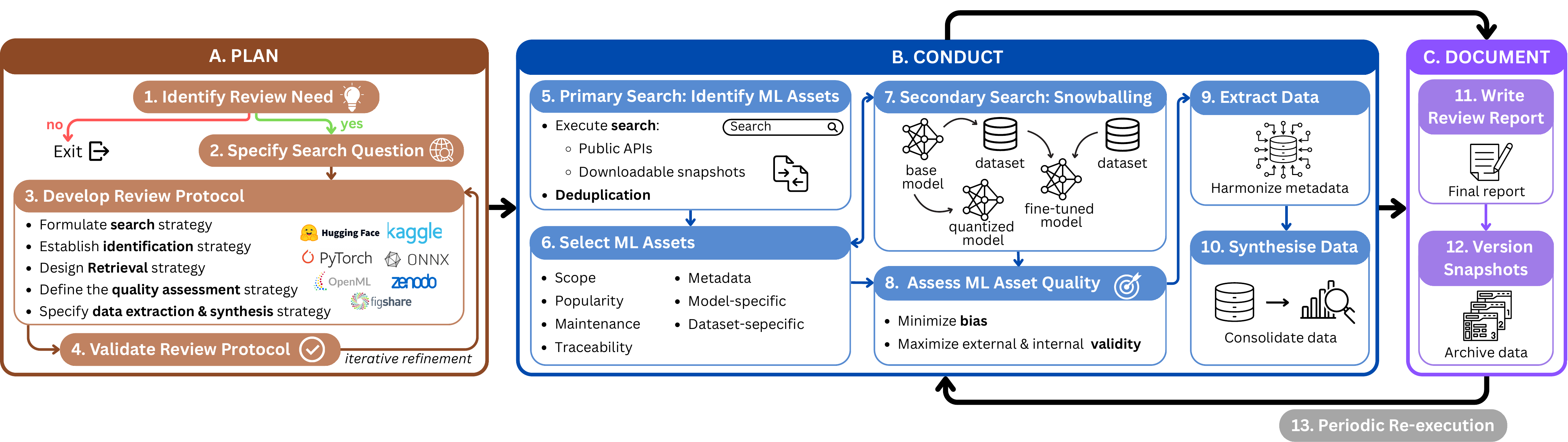}
    \caption{Systematic Review of ML Assets process.}
    \label{fig:schema}
\end{figure}

\subsection{Plan the Review}
Before conducting the review, it is necessary to assess its necessity, as this determines whether a systematic retrieval effort is justified. Once this need is established, the next step is to formulate the search questions that guide the review, ensuring that they are precise enough to delimit the ML asset space. This phase concludes with the definition and validation of the review protocol, which specifies the operational procedures, inclusion boundaries, and methodological choices that will structure the subsequent phases.

\textbf{\circled{1} Identify Review Need}. 
Given the rapid evolution and scale of AI registries, an ML asset review begins with a preliminary assessment to determine whether a systematic retrieval effort is warranted and not already addressed by prior studies. Researchers must examine limitations in existing asset collections, such as missing quality assessments, incomplete dependency mappings, or outdated evaluation results. This evidence is then used to justify the resources required for a rigorous systematic review.

\textbf{\circled{2} Specify Search Question}. 
The review process is grounded in a clearly articulated search question that can be derived from a broader Research Question (RQ). Considering ML assets, this step requires translating a study’s high-level objective into a specification of the ML asset type and its intended role within the study. Here, we need to clarify whether the intended retrieval target corresponds to a model, dataset, or benchmark, and how this ML asset is expected to support the research goal. 

For example, a broader RQ such as: ``How effective are ML approaches for automating code generation tasks?'' may be decomposed into multiple ML asset search questions:
\begin{itemize}
    \item ``Which \textit{models} are designed or adapted for source code generation?''
    \item ``Which \textit{datasets} are used to train ML models for code generation?''
    \item ``Which \textit{benchmarks} are used to evaluate the performance of code generation models?''
\end{itemize}
In this case, each search question operationalizes a specific aspect of the broader research intent by targeting distinct types of ML assets relevant to code generation tasks.

\textbf{\circled{3} Develop Review Protocol}. 
Establishing the protocol in advance is essential to reduce researcher bias and improve reproducibility. In this context, a pre-defined protocol documents all elements of the review together with the associated planning information.
Beyond the \textit{background}, which motivates the review, and the \textit{search questions} that guide the investigation, the protocol must define the following elements: 

\begin{itemize}
    \item \textit{Search strategy}: Specifies target registries while accounting for their structural heterogeneity and considering their primary focus. \textit{Model-centric} registries are primarily oriented toward hosting pre-trained models, including registries such as Model Zoo, PyTorch Hub, ONNX Model Zoo, NVIDIA NGC, and MATLAB Model Hub. In contrast, \textit{dataset-centric} registries focus on the reuse of datasets and include OpenML, Harvard Dataverse, Zenodo, and Figshare. Finally, \textit{mixed-purpose} registries such as Hugging Face and Kaggle support multiple ML asset types, including both models and datasets.

    \item \textit{Identification strategy}: Determines how registry data will be accessed, including public APIs, downloadable registry dumps, web scraping procedures, or hybrid approaches. Such decisions are important because registries differ in the amount and granularity of accessible data. For instance, popularity signals are not consistently exposed across registries. At the time of writing, Kaggle provides dataset-level download counts via its web interface and API, whereas equivalent signals for models are only available through the web interface. Registries also differ in identifier schemes, ranging from human-readable composite identifiers (e.g., \texttt{user\_id/asset\_id} in Hugging Face and Kaggle) to opaque numeric identifiers in their APIs (e.g., OpenML datasets identified by integer IDs). 
    
    \item \textit{Selection criteria}: Defines the inclusion and exclusion criteria for ML assets. These should be adapted to the characteristics of the target registries, as each registry may differ in structure and available data. They may include several dimensions, such as scope, traceability, and maintenance activity. Since registry information may be incomplete, the protocol should also specify how missing information will be handled.
    
    \item \textit{Quality assessment strategy}: Defines how ML assets will be evaluated in terms of verifiability of reported claims (e.g., performance metrics, training configurations), internal validity of the experimental evidence (sufficiency of methodological detail), and external validity of the ML assets.
    
    \item \textit{Data extraction and synthesis strategy}: Specifies which attributes and dependencies will be extracted from selected ML assets, as well as how these data will be synthesized. 
    
    \item \textit{Dissemination strategy}: Covers reporting guidelines, versioning practices, and the planned schedule for conducting and re-executing the review.
\end{itemize}

\textbf{\circled{4} Validate Review Protocol}.  
Evaluating the protocol before full-scale execution is critical for discovering mistakes. Protocol validation should confirm that the proposed search strategy retrieves relevant ML assets across the selected registries. In practice, validation may involve executing pilot searches over a representative subset, inspecting the retrieved data, and verifying whether the resulting set adequately reflects the intended review scope. Based on these results, the protocol may be iteratively refined.

\subsection{Conduct the Review}
The second phase of the review operationalizes the protocol through systematic identification, filtering, and analysis of ML assets. 
During this phase, researchers execute the search strategy by identifying ML assets, and applying the inclusion, exclusion, and quality assessment criteria established in the protocol. In addition, snowballing should be employed to identify further relevant ML assets. Once the final set of ML assets has been selected, researchers can perform data extraction and synthesize the collected evidence. 

Given the potentially large number of ML assets retrieved from AI registries such as Hugging Face, which hosts 2.9M models and 1M datasets as of May 2026, the execution of this phase may require support from automated techniques. Recent studies explore the use of LLMs to support empirical workflows in mining software repositories \cite{gomes2026llm, romero2026rise}, as well as structured prompting and validation frameworks such as PRIMES \cite{de2025framework} and empirical guidelines for SE studies involving LLMs \cite{baltes2025guidelines}. These approaches emphasize combining LLM-assistance with rule-based filtering and human validation to improve scalability while preserving reliability and mitigating risks such as hallucinations.

\textbf{\circled{5} Primary Search: Identify ML Assets}. 
The primary search aims to construct an initial set of candidate ML assets. Each search question specified in the protocol is translated into one or more search queries by identifying core concepts (e.g., ML asset type, task) and expanding them with synonyms and registry-specific terminology. These queries are not executed against a uniform bibliographic database but across heterogeneous AI registries that expose different retrieval mechanisms, including free-text search and metadata filters. As a result, search strings are operationalized as registry-specific queries, combining keywords with filtering tags supported by each registry and adapted to their indexing mechanisms.

Search execution is performed using the available interfaces of each registry. Access to registry data can be achieved either through public APIs or via downloadable snapshots. APIs enable fine-grained querying over metadata fields and textual descriptions, while static dumps provide a complete view of the registry at a given point in time. In API-supported registries, queries are issued through structured endpoints, whereas in registries without queryable APIs, search is performed through web interfaces or downloaded snapshots. A key challenge in API-based retrieval is that APIs often expose incomplete data, omitting information that is only available on the registry’s web interface. To address this limitation, primary search may rely on hybrid retrieval strategies that combine structured API queries with web scraping of ML asset pages to recover missing information directly from the source. However, researchers must account for registry constraints such as rate limits, which bound the volume and frequency of requests, thereby limiting large-scale or highly iterative extraction workflows. As a result, primary search is often conducted in an incremental and iterative manner.

After identifying candidates, duplicates may arise due to mirrored repositories or cross-registry republishing. To address this, deduplication must be applied based on information available in the documentation. ML assets with highly similar documentation are considered copies, and timestamps are used to distinguish between original and replicas \cite{gonzalez2025cataloguing}.

\textbf{\circled{6} Select ML Assets}.
Following the identification of candidate primary ML assets, the next step consists of selecting those that satisfy the inclusion and exclusion criteria defined in the review protocol. To perform this task, reviewers must examine the information associated with each ML asset in the registry, since the relevance of an ML asset cannot be determined from its identifier alone. In AI registries, a natural starting point for this selection process is the inspection of model and dataset cards \cite{mitchell2019model, gebru2021datasheets}. These combine structured metadata, commonly represented through YAML fields (e.g., license type, library), and unstructured natural language descriptions written in Markdown that provide contextual information, including intended use and limitations. 
As summarized in Table \ref{tab:selection-dimensions}, the selection process comprises multiple complementary dimensions to capture ML asset suitability. It is worth noting that these dimensions are conditioned by the signals available in each registry, since not all registries expose the same information. 

The extent to which an ML asset matches the review is captured through \textit{scope} alignment. This is operationalized by examining the tasks explicitly associated with the ML asset in the registry, such as `text classification', as well as related task annotations provided in its card.

In addition, adoption within the ecosystem is approximated through \textit{popularity}. This is derived from signals such as download counts, likes, or upvotes, when available. While highly popular ML assets may indicate broad community validation and reuse, this measure must be interpreted cautiously due to temporal bias, since recently published ML assets may not have had sufficient time to accumulate engagement despite being relevant. 

Similarly, the level of ongoing support is reflected through \textit{maintenance} activity, which captures how actively an ML asset is updated over time. This is inferred from indicators such as the date of the last update, commit history, and contributor engagement. ML assets with recent and sustained maintenance are generally preferred, as they are more likely to remain compatible with evolving frameworks and dependencies.

Another relevant aspect concerns \textit{traceability}, which describes the ability to connect an ML asset to related resources, such as parent/derived variants, datasets used in downstream or upstream pipelines, or publications associated with a given ML asset. This information is typically available through dedicated sections in the cards, or as unstructured references embedded in the description. For example, a model may detail the base model from which it was derived, fine-tuned, or adapted versions of it, as well as the training datasets used. 

Beyond these aspects, the quality of documentation is reflected through \textit{metadata} completeness, which captures how thoroughly an ML asset is described to support informed assessment and reuse. This is evaluated based on the presence of fields, such as license information, library dependencies, and ML asset size. When such information is missing or incomplete, the ability to verify eligibility is significantly reduced.

Finally, technical characteristics are considered through two complementary perspectives. On the one hand, \textit{model-specific} characteristics capture properties of trained models, including training data, evaluation benchmarks, and reported performance metrics. On the other hand, \textit{dataset-specific} ones describe properties of the underlying data, such as modality and format. 

\begin{table}[!htb]
\centering
\small
\begin{tabular}{p{2.3cm} p{5.3cm} p{5cm}}
\toprule
\textbf{Dimension} & \textbf{Purpose} & \textbf{Examples} \\
\midrule
Scope & Determine relevance to the review & ML task, application domain \\
Popularity & Estimate usage and visibility & Likes, downloads, upvotes \\
Maintenance & Evaluate evolution and support & Recent updates, contributors \\
Traceability & Identify dependency relationships & Parent/derived ML assets \\
Metadata & Assess documentation sufficiency & License, library, size \\
Model-Specific & Capture model characteristics & Training dataset, evaluation data\\
Dataset-Specific & Characterize data properties & Modalities, format \\
\bottomrule
\end{tabular}
\caption{Selection dimensions used in the systematic review of ML assets.}
\label{tab:selection-dimensions}
\end{table}

\textbf{\circled{7} Secondary Search: Snowballing}. 
After identifying and selecting primary ML assets, researchers can extend the retrieval process by applying snowballing strategies \cite{wohlin2014guidelines} adapted to AI registries. Here, ML assets form interconnected ecosystems in which models and datasets are often linked through explicit metadata or implicit references. 

In this context, \textit{backward snowballing} can be performed by inspecting the provenance information of a selected ML asset, i.e., identifying the resources from which it originates. 
For models, this involves tracing the base model and the training datasets. For instance, a code-generation model may indicate that it was fine-tuned from a general-purpose model and trained on curated code corpora. In Hugging Face, this is explicitly supported through structured metadata fields such as \texttt{base\_model}, which links a model to its parent models, and \texttt{datasets}, which indicates training datasets. For datasets, backward snowballing involves identifying their origin sources, such as upstream datasets from which they were derived.

Conversely, \textit{forward snowballing} can be conducted by identifying downstream ML assets that reuse a given model or dataset. For models, this includes fine-tuned versions, adapters (e.g., LoRA or PEFT variants), quantized models, or merged architectures derived from a base model. For datasets, forward snowballing involves identifying downstream ML assets that are built on top of a given dataset, such as models trained or fine-tuned using that dataset, or benchmarks that incorporate it as part of their evaluation pipeline. 

More broadly, this process reflects that ML assets do not exist in isolation but form interconnected ecosystems of reuse. In their work, Yang et al. \cite{yang2025ecosystem} operationalize this idea by manually performing snowballing over a selected set of LLMs for code, where they identified training datasets and related models. However, such relationships are not always explicitly encoded in structured metadata and often appear only in unstructured descriptions. As a result, snowballing may require inspection of textual fields to recover implicit dependencies.

Lastly, all newly identified ML assets through snowballing are subsequently subjected to the same inclusion and exclusion criteria defined in the review protocol, and are only incorporated into the final set after passing the selection step.

\textbf{\circled{8} Assess ML Asset Quality}. 
Once the final set of ML assets has been identified and expanded through snowballing, a quality assessment is performed to evaluate its reliability. In contrast to selection, which determines whether an ML asset is relevant to the review scope, this step focuses on the strength of the information that supports its use in subsequent steps. Following established systematic review principles, quality is understood in terms of the extent to which bias is minimized and internal and external validity are maximized. In this context, bias relates to uncertainty introduced by inconsistently reported information, while validity relates to the applicability of the evidence provided.

The assessment considers the extent to which claims associated with an ML asset can be independently verified, including reported performance, training data composition, or intended use, and whether these elements are supported by external sources such as publications. This relates to the presence of potential bias in registry-reported information. It also evaluates the credibility of the experimental evidence provided, focusing on whether reported results are accompanied by sufficient methodological detail to support internal validity, including the clarity of evaluation protocols, the definition of benchmarks, and the availability of configuration information that contextualizes performance claims. A further aspect concerns the applicability of the evidence outside the original reporting context, particularly in situations where evaluation settings are ambiguously defined, where training and evaluation boundaries are not clearly specified, or where preprocessing and data construction steps are insufficiently documented, thereby limiting external validity during synthesis.

\textbf{\circled{9} Extract Data}.
Once the final set of ML assets has been selected, relevant attributes are collected to answer the initial research questions and enable subsequent analysis.

Given the variability of metadata across registries, extraction involves harmonizing representations of equivalent attributes and extracting relevant unstructured information. This includes aligning different naming conventions, resolving inconsistencies in identifiers, and mapping registry-specific fields to a unified schema defined in the review protocol. In addition, relevant information is also retrieved from descriptive text associated with each ML asset. Model cards and dataset cards often contain details such as intended use, training procedures, limitations, and known biases. For instance, a dataset card may describe filtering steps applied to remove low-quality or duplicated samples, while a model card may provide information about architecture choices or fine-tuning strategies. These details are extracted through manual inspection or automated parsing techniques, depending on the review setup.

When certain attributes are not available, their absence is explicitly recorded. This is particularly relevant in cases where key information, such as training data provenance or evaluation benchmarks, is omitted. In such situations, the missingness itself becomes an analytical signal that may inform the interpretation of results in subsequent phases.

\textbf{\circled{10} Synthesise Data}. 
The final step of the conducting phase aggregates, normalizes, and categorizes the extracted data to answer the initial search questions. 
As AI registries often yield an unmanageably large corpus of valid ML assets, a stratified sampling approach can be applied to extract a representative subset \cite{castano2024lessons}. This strategy ensures that the synthesis is not biased toward dominant ML assets, but remains inclusive of the ecosystem's variety.

\subsection{Document the Review}
The last phase of the review involves documenting it. Specifically, it focuses on ensuring transparency, reproducibility, and longitudinal traceability of the retrieval process. It records search strategies, selection decisions, and extracted data. In addition, it supports iterative updates to account for the dynamic nature of AI registries, enabling continuous refinement and replication of the review process over time.

\textbf{\circled{11} Write Review Report}: 
A structured report captures the full review process, including the rationale behind methodological decisions, the applied search strategy, and the criteria used for ML asset selection. It also records deviations from the initial protocol and provides a clear mapping between search questions and retrieved ML assets. Following established reporting guidelines such as PRISMA \cite{page2021updating}, the goal is to ensure that the review can be critically assessed, reproduced, and extended by other researchers.

\textbf{\circled{12} Version Snapshots}:
Version-controlled snapshots of the review must be maintained to preserve the state of AI registries at specific points in time. Unlike traditional literature corpora, ML assets are continuously updated, modified, or deprecated, which introduces a significant threat to reproducibility if only static records are retained. For instance, model weights may be updated, datasets may be expanded with new samples or corrected labels, and metadata such as performance metrics, licenses, or documentation may be revised as communities refine or correct earlier releases. For this reason, review reporting cannot be treated as a single static extraction, but as a sequence of reproducible snapshots. Each snapshot represents a frozen view of the unified database, enabling consistent analysis and comparison across different time points.

\subsection{Need of Iteration}
\textbf{\circled{13} Periodic Re-execution}: Given the dynamic and continuously evolving nature of AI registries, a one-off execution of the review process is insufficient to maintain up-to-date results. Therefore, we define an iterative loop over the \textit{Conduct Review} and \textit{Document Review} phases, while the \textit{Plan Review} phase remains fixed after its validation to ensure consistency across iterations. The inclusion of new AI registries or any change to the protocol is excluded from the iterative mechanism, as it alters the study definition and affects comparability. Such cases require a new review instance rather than a new iteration.

Within this loop, the \textit{Conduct Review} phase is periodically re-executed to refresh retrieval, filtering, and analysis of ML assets as registries evolve. This re-execution captures newly added or modified ML assets within those registries. 
Then, the \textit{Document Review} phase is updated to record new versioned snapshots and preserve traceability of changes over time. This iterative mechanism reduces manual effort, mitigates drift in ML asset collections, and ensures that the review remains aligned with the current state of the ecosystem within a fixed methodological scope.

\section{Discussion}
Systematic retrieval of ML assets sits at the intersection of established empirical methodologies and emerging challenges introduced by AI registries. Established empirical standards (e.g., ACM/SIGSOFT guidelines) are grounded in document-centric assumptions, and search strategies such as snowballing \cite{wohlin2014guidelines} extend retrieval by exploiting citation relationships. In parallel, mining software repositories research has extended empirical analysis toward software artifacts in platforms such as GitHub \cite{dabic2021sampling}. However, AI registries introduce a setting in which metadata is sparse, inconsistent, and heterogeneous \cite{jiang2025see}. These characteristics make ML assets difficult to fully capture using either traditional systematic reviews of scientific literature or repository mining approaches. Therefore, we position the proposed framework as an adaptation of existing evidence synthesis principles that highlights ML assets as first-class analytical units. The key distinction lies in shifting the unit of analysis from documents or repositories to ML assets as interconnected artifacts.

This perspective aligns with Multivocal Literature Reviews (MLRs) \cite{garousi2019guidelines}, which incorporate grey literature (e.g., preprints, e-prints) alongside published literature (i.e., journal papers, conference proceedings, and books). AI registries introduce a comparable challenge, since relevant information about ML assets (e.g., evaluation signals) is not always peer-reviewed. However, while MLRs primarily integrate textual evidence, this framework extends multivocality toward ML assets.

From a broader empirical SE perspective, the proposed framework is also related to data strategies in empirical data-driven research \cite{storey2020software}. In particular, it resonates with the idea that the validity of empirical findings depends not only on analysis techniques but also on how data is selected, sampled, and curated from large and evolving AI ecosystems. In this sense, ML asset retrieval can be seen as a form of structured sampling over AI registries, where representativeness, traceability, and bias control become central concerns.

Finally, the lack of traceable ML asset retrieval directly impacts reproducibility in empirical SE. As open science policies become more important in venues such as ESEM or ICSE, transparency must extend beyond code to include the full process of artifact selection. This includes how models and datasets are identified, filtered, and justified. By treating ML asset retrieval as a versioned process, the framework operationalizes reproducibility at the level of ecosystem exploration, turning what is typically an ad hoc methodological step into a verifiable and repeatable component of empirical research.

\section{Conclusion \& Future Work}
This emerging results paper proposes a framework for the systematic review of ML assets in AI registries, addressing the lack of structured practices in state-of-the-art studies. By adapting established evidence synthesis principles and lessons learned from pioneering studies, the framework accounts for the heterogeneous, dynamic, and weakly standardized nature of AI ecosystems. It provides a structured process for planning, conducting, and documenting ML asset retrieval, enabling more transparent selection compared to ad hoc practices. Therefore, the framework offers a foundation for more systematic ML asset discovery.

Future work will focus on further formalizing the proposed methodology into a more operational framework, including mapping its steps against existing literature to identify established, partially addressed, and implicit practices. We also aim to investigate how the framework can be empirically validated, including what constitutes evidence of effectiveness for ML asset retrieval, how consistently the approach can be applied across heterogeneous AI registries, and how it compares against ad hoc selection practices commonly used in empirical studies. At this stage, the approach is grounded in prior systematic review methodologies and empirical studies of AI registries, and should be interpreted as an empirically informed design rather than a fully validated method. Finally, we envision the construction of a knowledge graph of the AI ecosystem where models, datasets, and their relationships are explicitly represented. This would enable ecosystem-level modeling and make retrieval strategies such as snowballing more systematic and straightforward.

\section{Data Availability}
This paper presents a conceptual and methodological framework. No datasets, code, or other research artifacts were collected, generated or analyzed as part of this article.

\bibliography{references}

\end{document}